\def\year{2022}\relax
\documentclass[letterpaper, 10 pt, conference]{ieeeconf}
\IEEEoverridecommandlockouts                              % This command is only needed if
\usepackage{times}  % DO NOT CHANGE THIS
\usepackage{helvet}  % DO NOT CHANGE THIS
\usepackage{courier}  % DO NOT CHANGE THIS
\usepackage[hyphens]{url}  % DO NOT CHANGE THIS
\usepackage{graphicx} % DO NOT CHANGE THIS
\usepackage{cite}
\usepackage{caption} % DO NOT CHANGE THIS AND DO NOT ADD ANY OPTIONS TO IT
\DeclareCaptionStyle{ruled}{labelfont=normalfont,labelsep=colon,strut=off} % DO NOT CHANGE THIS
\usepackage{algorithm}
\usepackage{moreverb}

\newif\ifanonym
\anonymfalse

\usepackage{newfloat}
\usepackage{listings}
\floatstyle{ruled}
\newfloat{listing}{tb}{lst}{}
\floatname{listing}{Listing}
\ifanonym
\else
\fi

\usepackage{cite}
\usepackage{caption} % DO NOT CHANGE THIS AND DO NOT ADD ANY OPTIONS TO IT
\DeclareCaptionStyle{ruled}{labelfont=normalfont,labelsep=colon,strut=off} % DO NOT CHANGE THIS
\usepackage{algorithm}
\usepackage{moreverb}

\newif\ifanonym
\anonymfalse

\usepackage{newfloat}
\usepackage{listings}
\floatstyle{ruled}
\newfloat{listing}{tb}{lst}{}
\floatname{listing}{Listing}
\ifanonym
\else
\fi

\usepackage[compact]{titlesec}
\titleformat{\subsubsection}[runin]
{\itshape}{\thesubsubsection}{0em}{}

\title{Constructing Behavior Trees from Temporal Plans for
Robotic Applications\\[-5mm]%
\ifanonym
\else
\fi
}
\ifanonym
  \author{ICAPS-23 submission 4774}
\else
\author {
    Josh Zapf\textsuperscript{\textrm{1}},
    Marco Roveri\textsuperscript{\textrm{2}},
    Francisco Martín\textsuperscript{\textrm{3}},
    Juan Carlos Manzanares\textsuperscript{\textrm{3}}
        \thanks{\textsuperscript{\textrm{1}} NIWC Pacific, San Diego, USA - joshua.j.zapf.civ@us.navy.mil}
        \thanks{\textsuperscript{\textrm{2}} University of Trento, Trento, Italy - marco.roveri@unitn.it}
    \thanks{\textsuperscript{\textrm{3}} Intelligent Robotics Lab, Rey Juan Carlos University, Fuenlabrada, Spain - francisco.rico@urjc.es, juancarlos.serrano@urjc.es}
}
\fi

\usepackage[noend]{algpseudocode}
\usepackage{amsmath,amsfonts}
\usepackage[disable]{todonotes}
\presetkeys{todonotes}{inline}{}

\let\labelindent\relax
\usepackage[inline]{enumitem}
\newlist{algitem}{itemize*}{4}
\setlist[algitem]{label=}

\newcommand\CONDITION[2]%
  {\begin{tabular}[t]{@{}l@{}l@{}}
     #1&#2
   \end{tabular}%
  }

\newtheorem{theorem}{Theorem}

\newtheorem{lemma}{Lemma}
\newtheorem{definition}{Definition}

\newcommand{\astart}[1]{\ensuremath{{#1}_{\vdash}}}
\newcommand{\aend}[1]{\ensuremath{{#1}_{\dashv}}}
\newcommand{\ainv}[1]{\ensuremath{{#1}_{\vdash\dashv}}}

\begin{document}

\maketitle

\begin{abstract}
Executing temporal plans in the real and open world requires adapting to uncertainty both in the environment and in the plan actions.
A plan executor must therefore be flexible to dispatch actions based on the actual execution conditions.
In general, this involves considering both event and time-based constraints between the actions in the plan.
A simple temporal network (STN) is a convenient framework for specifying the constraints between actions in the plan.
Likewise, a behavior tree (BT) is a convenient framework for controlling the execution flow of the actions in the plan.
The principle contributions of this paper are
\begin{enumerate*}[label=\roman*)]
\item an algorithm for transforming a plan into an STN, and
\item an algorithm for transforming an STN into a BT.
\end{enumerate*}
When combined, these algorithms define a systematic approach for executing total-order (time-triggered) plans in robots operating in the real world.
Our approach is based on creating a graph describing a deordered (state-triggered) plan and then creating a BT representing a partial-order (determined at runtime) plan.
This approach ensures the correct execution of plans, including those with required concurrency.
We demonstrate the validity of our approach within the PlanSys2 framework on real robots.
\end{abstract}

\section{Introduction}

%Robotic agents to operate in real world environments need to execute actions with a duration, that are possibly executed in concurrently (e.g., while moving set up a tool for immediately start an experimet when at destination), and with temporal constraints among the actions.
%
We are interested in executing temporal plans with concurrent actions on real robots in the real wold.
In its most basic form, a temporal plan is one in which the actions of the plan have durations and there exist temporal (along with event-based) constraints between the actions.
Such plans can be computed offline by state-of-the-art temporal planners~\cite{ICAPS101421} starting from a planning problem formulated in PDDL 2.1~\cite{fox2003pddl2}.
Executing temporal plans in the real world is complicated both by uncertainty in the environment and in the actions themselves.
%
% While the plans are created offline, there are several factors to consider when executing them~\cite{Munson1971RobotPE}.
%
% First of all, although the specification of an action assumes a duration used to calculate the plan, in reality, the duration of an action can vary considerably.
%
% Imagine a navigation action of a robot.
%
% Depending on the calculated route to the destination, and the difficulties that the robot may encounter along the way, the duration will always be different from what we can declare in a PDDL domain.
%
% Secondly, in PDDL, a closed world is assumed where the facts only vary due to the planned actions.
%
% Finally, while the plans generated by a PDDL planner indicate at what instant of time each action must begin and end, in reality, an action begins when the action on which it depends applies the effects that satisfy its preconditions.
%
% Therefore, a robot that operates in the real world cannot count on these underlying assumptions.
%
A robotic agent must therefore be capable of representing plans in a flexible manner.
%
%Therefore, a robotic agent to execute such plans need to be equipped with a plan executor able to represent such plans, and to consider both event and time-based constraints between the actions in the plan.
%
A fairly efficient approach is to encode temporal plans as graphs, where the nodes of the graph are the actions and the arcs define the order between the actions.
This representation is made even more complete if we divide the durative actions into two nodes: their beginning and end~\cite{strathprints77023}.
This type of planning graph can be executed in a framework such as ROSPLAN~\cite{10.5555/3038662.3038708} or the ROS 2 Planning System (PlanSys2)~\cite{PlanSys2}.
Additionally, behavior trees (BTs) have been proposed for plan execution~\cite{DBLP:phd/basesearch/Colledanchise17,DBLP:conf/icra/MarzinottoCSO14} and provide a convenient means for describing the switchings between a finite set of tasks in a modular fashion.
%
%and are becoming a de-facto standard in many robotic applications~\cite{DBLP:conf/iros/SafronovCN20}.
%
%They indeed offer a flexible formalism for creating complex behaviors from simpler ones and can adequately represent complex PDDL plans, including the causal constraints between the actions in the plan, the evaluation of preconditions, and the application of effects with correct timing.
%
PlanSys2 first proposed the encoding of a temporal plan into a BT~\cite{DBLP:conf/atal/RicoMELO21}.
However, this approach did not capture the full semantics of PDDL 2.1 and could not handle the case where all or part of an action is constrained to fall between the start and end of another.
To construct a robust and correct execution, perfectly representing the order between the actions of a plan, the encoding must take into account the relationships between the start and end requirements and effects of the actions in the plan.
To this end, we have developed a theoretical framework and related algorithms for transforming a general temporal PDDL plan into
\begin{enumerate*}[label=\roman*)]
\item an STN and
\item a BT
\end{enumerate*}
for robust execution on real robots that preserves the semantics of the original temporal plan.
This allows our system to be used for the execution of plans solving a wider range for problems, including those with required concurrency~\cite{DBLP:conf/ijcai/CushingKMW07}.
We have implemented the proposed framework within the PlanSys2 framework, and we demonstrated its validity by experimenting both on artificial and real-world robotic problems exhibiting different degrees of concurrency.
The results show that the proposed encoding correctly implements the semantics of the PDDL temporal plans and results in a robust and flexible framework for executing temporal plans in robotic applications.

This paper is organized as follows.
In Section~\ref{sec:rw}, we summarize the related work.
In Section~\ref{sec:bg}, we summarize the background.
The core of the paper is Section~\ref{sec:bt_encoding}.
Section~\ref{sec:validation} reports the results of the experimental validation and discusses possible limitations.
Finally, in Section~\ref{sec:conclusions}, we draw our final conclusions and outline some future directions.

\section{Related Work}
\label{sec:rw}

% \todo[inline]{
%   Shall we keep it here or shall we move it later highlighting what are the diferences with our approach?
%   \begin{itemize}
% \item The works on PlanSys2 by Francisco
% \item The works on ROSPLAN (e.g. the IntEx papers)
% \item Other works on BT for planning?
% \end{itemize}
% }

In order to make robotic applications flexible and adaptable to changes and contingencies, there is a need for automatic plan generation, where a plan can be thought of as a course or schedule of possibly parallel elementary actions needed to achieve the goal~\cite{DBLP:journals/arobots/NourbakhshG96}.
We refer the reader to~\cite{DBLP:journals/corr/abs-2101-01964} for a thorough discussion on the use of planning in Robotics.

The reference framework for the automatic generation of behaviors for a robotic agent in ROS 1~\cite{ROS} is ROSPlan~\cite{10.5555/3038662.3038708}.
In a nutshell, ROSPlan dispatches the execution of the actions of a time triggered plan such that the order of action execution is causally complete (each event’s conditions are satisfied) and temporally consistent with respect to the simple temporal network (STN) induced by the temporal plan.
The action dispatch consists of activating a software component responsible for the semantics of the action at the correct time as defined by the STN.
See~\cite{10.5555/3038662.3038708} for further details.

% The approach adopted in ROSPlan to dispatch the actions consists of taking the time triggered plan generated by a temporal planner, create a corresponding simple temporal network, and use this structure to dispatch the actions.

Behavior trees (BTs) have been proposed as well founded mathematical models for plan execution~\cite{DBLP:phd/basesearch/Colledanchise17,DBLP:conf/icra/MarzinottoCSO14} and are becoming a de-facto standard in many robotic applications~\cite{DBLP:conf/iros/SafronovCN20}.
They indeed offer a flexible formalism for creating complex behaviors from simpler ones and can adequately represent complex PDDL plans, including the causal constraints between actions, the evaluation of preconditions, and the application of effects.

A few approaches have been proposed for automatically generating BTs from high-level plans.
In~\cite{8901959} a continuous updating approach was proposed based on the Hierarchical task and motion Planning in the Now (HPN)~\cite{DBLP:conf/aaai/KaelblingL10} algorithm.
In \cite{DBLP:conf/atal/RicoMELO21} and its extended version~\cite{DBLP:journals/corr/abs-2101-01964} an approach was proposed that goes beyond a domain-dependent scenario associated with the hierarchical formalization.
Their algorithm leverages the structure of the plan to build a BT that optimizes the execution time by explicitly producing parallel actions even in case where the original plan is purely sequential.
For the conversion, they
\begin{enumerate*}[label=\roman*)]
\item introduce a new kind of BT node to wait on the execution of an action either running in parallel or
not yet executed, which may appear in any other part of the tree, and
\item use a singleton BT to encode an action, which allows for the execution of the same action from various points in the tree.
\end{enumerate*}

The approach taken by ~\cite{DBLP:journals/corr/abs-2101-01964} has been integrated into the PlanSys2~\cite{PlanSys2} planning framework for ROS 2~\cite{doi:10.1126/scirobotics.abm6074}.
This approach has proven to be efficient for many practical applications.
However, the generated BTs do not capture precisely the semantics of complex time triggered plans.
Indeed, for some plans with complex interactions between the start and end effects of two or more actions, the generated BTs result in the invalid application of action effects.
That is, some of the actions are deployed at the wrong times.
Our work builds on and extends this preliminary work.
Starting from the time triggered plan, we
\begin{enumerate*}[label=\roman*)]
\item convert the plan into a simple temporal network (STN), where each node corresponds to a start or end snap action in the induced simple plan,
\item optionally convert the STN into a dispatchable one~\cite{DBLP:conf/kr/MuscettolaMT98},
\item convert each node in the STN to a singleton BT, and
\item connect these into a single BT representing the entire plan.
\end{enumerate*}
This process guarantees that the generated BT will be correct by construction w.r.t. the initial time triggered plan.

\section{Background}
\label{sec:bg}

In this section we summarize the most relevant background concepts used through the whole paper.

%\todo[inline]{%
% \begin{itemize}
% %\item Planning - Done
% \item Behavior Tree (we need to introduce them)
% %\item A short introduction to PlanSys2?
% \item ...
% \end{itemize}
%This section is currently too long. It can be fine for an extended version to publish on arxiv, but it is definitely too long for the conference. The limit is 8 pages + 1 page of references
%}

\subsection{Temporal Planning Problem}

We define, following \cite{DBLP:books/daglib/0014222}, a \emph{(STRIPS) classical planning problem} as a tuple $CP = (F, O, I, R_g)$, where $F$ is the set of fluents, $O$ is a finite set of actions, $I \subseteq F$ is the initial state, and $R_g \subseteq F$ is a goal condition.
A literal is either a fluent or its negation.
Every action $a \in O$ is defined by two set of literals: the preconditions, written $pre(a)$, and the effects, $eff(a)$.
A \emph{classical plan} $\pi = (a_1, \cdots, a_n)$ is a sequence of actions. A classical plan \emph{is valid} if and only if it is executable from the initial state and terminates in a state fulfilling $R_g$.
%
%This formulation follows the one presented in \cite{DBLP:books/daglib/0014222}.

We define a \emph{temporal planning problem} as a tuple $TP = (F, DA, I, R_g)$, with $F$, $I$ and $R_g$ defined as in the classical planning problem, and with $DA$ being a set of \emph{durative actions}.
Following \cite{DBLP:conf/ijcai/CushingKMW07}, a durative action $a
\in DA$ is given by
\begin{enumerate*}[label=\roman*)]
\item two classical planning actions $\astart{a}$ and $\aend{a}$;
\item an overall condition $overall(a)$ expressed as a set of literals;
\item and a minimum $\delta_{min}(a)\in \mathbb{R}^+$ and maximum $\delta_{max}(a) \in \mathbb{R}^+$ duration, with $\delta_{min}(a) \le \delta_{max}(a)$.
\end{enumerate*}
A \emph{temporal plan} $\pi = \{tta_1, \cdots, tta_n\}$ is a set of time triggered temporal actions, where each $tta_i$ is a tuple $\langle t_i, a_i, d_i \rangle$ where $t_i \in \mathbb{R}^+$ is the starting time, $a_i \in DA$ is a temporal action, and $d_i \in [\delta_{min}(a_i), \delta_{max}(a_i)]$ is the action duration.
We say that $\pi$ is a \emph{valid temporal plan} if and only if it can be
simulated (i.e., starting from the initial model we apply each time
triggered action $tta_i = \langle t_i, a_i, d_i \rangle \in \pi$ at time $t_i$ with duration $d_i$), and at the end of the simulation we obtain a state fulfilling the goal condition.
For lack of space we refer the reader to \cite{pddltwoone} for a thorough discussion on the semantics and definition of a \emph{valid temporal plan}.
Many approaches to temporal planning transform each durative action $a \in DA$ into three \emph{snap actions}: $\astart{a}$ and $\aend{a}$ that contain the preconditions and effects of the start and end of $a$, $\ainv{a}$ used during the search to enforce the overall condition $overall(a)$ between $\astart{a}$ and $\aend{a}$, and additional fluents modified by $\astart{a}$, $\ainv{a}$, and $\aend{a}$ to enforce their temporal ordering (see e.g.~\cite{DBLP:journals/ai/ColesFHLS09} for further details).

\subsubsection{Induced Simple Plan.}

Given a temporal plan, one can compute an \textit{induced simple (temporal) plan}, which can be thought of as a sequence of classical planning actions with associated time instants.
We leverage the induced simple plan for constructing the planning graph discussed later on, which is necessary to build the corresponding behavior tree.
For the purposes of this paper, we assume the given temporal plan is valid and do not address the validity problem.

\begin{definition}
\label{def: induced simple plan}
Let $\pi = \{\langle t_1, a_1, d_1 \rangle , \cdots, \langle t_n, a_n, d_n \rangle\}$ be a temporal plan. The set of \textbf{happening time points} is
\begin{align*}
H(\pi) = \{t | \langle t, a, d \rangle \in \pi\} \cup \{t+d | \langle t, a, d \rangle \in \pi\}.
\end{align*}
Furthermore, let $t_1,\ldots,t_k$ be the points in $HT(\pi)$ sorted in increasing order.
The \emph{induced simple plan}, $S$, is a multiset of timed instantaneous actions which includes
\begin{itemize}
\item $t:\astart{a}$ and $t+d:\aend{a}$ for each timed action occurrence $\langle t, a, d \rangle \in \pi$; and
\item $\left(\frac{t_i+t_{i+1}}{2}\right):\ainv{a}$ for each timed action occurrence $\langle t, a, d \rangle \in \pi$ and $t_i,t_{i+1} \in HT(\pi)$ such that $t \leq t_i$ and $t_{i+1} \leq t+d$.
\end{itemize}
\end{definition}

Each step in the induced simple plan corresponds to an instantaneous action derived from a durative action in the original plan.
For each durative action, there will be one start action $\astart{a}$, one end action $\aend{a}$, and zero to many overall actions $\ainv{a}$.
The number of overall actions $\ainv{a}$ depends on whether an $overall(a)$ condition is present in the original action and the number of happening points that fall between the start and end of the action.
Note that, while the $\ainv{a}$ actions are needed to check the validity of over all conditions, they do not correspond to a change in the state (they have the $overall(a)$ condition as a precondition, and no effect).
See e.g.~\cite{DBLP:journals/ai/ColesFHLS09} for further details.

\subsubsection{Causal Links.}

A causal link is a pair $(i,j)$, where $i$ and $j$ are steps in the induced simple plan $S$, such that $i \prec j$.
A causal link can be expressed as
%
%\begin{align}
$[\langle t, a, d \rangle^i,v] \to [\langle t, a, d \rangle^j,w]$,
%\end{align}
%
where $v,w \in \{\textsc{start},\textsc{end}\}.$

There are two types of causal links.
The first type occurs when an effect of one action satisfies a precondition of another.
The second type occurs when an effect of one action threatens, i.e., invalidates, a precondition or effect of another.
Given a valid plan, it is possible to find all causal links between the actions in the plan.

%To gain some intuition on threatening causal links, consider the following plan.
%%
%\begin{verbatim}
%0.000: (move r l1 l2)  [3.000]
%3.000: (spin r l2)  [1.000]
%4.000: (move r l2 l3)  [2.000]
%\end{verbatim}
%%
%Now, suppose \texttt{(is\_at r l2)} is an overall condition of \texttt{(spin r l2)} and a start condition of \texttt{(move r l2 l3)}.
%%
%Furthermore, suppose \texttt{(move r l2 l3)} has the start effect \texttt{(not (robot\_at r l2))}.
%%
%Thus, while \texttt{(move r l1 l2)} satisfies the conditions of the latter two actions, the latter two actions cannot run in parallel, since the start effect of \texttt{(move r l2 l3)} threatens the overall condition of \texttt{(spin r l2)}.

\subsubsection{Temporal Consistency.}

Temporal consistency refers to the valid scheduling of actions in a plan such that the preconditions of each action are satisfied.
In some cases, this implies that the start and end of a durative action needs to straddle one or more other actions.
A common example of this is when one action must be completed within the interval defined another.
In such instances, it is often necessary to offset the time at which an action starts relative to the start or end of another action.

%Consider the example shown in Figure \ref{fig: action coordination}.
%%
%Suppose action $a$ has a start effect $p$ that satisfies a start condition of action $b$ such that $b$ must start after $a$.
%%
%Suppose also that $a$ has an end effect $q$ that satisfies an end condition of $b$ such that $b$ must end after $a$.
%%
%That is, $b$ must straddle the end of $a$.
%%
%If the duration of $b$ is less than the duration of $a$, $b$ cannot start immediately after $a$.
%%
%Instead, $b$ should start some time after $a$ to ensure that it overlaps with the end of $b$.

%This example demonstrates the need to coordinate the action start times in order to maintain temporal consistency.
%%
%In general, it is not possible to schedule the end of an action based only on the end preconditions.
%%
%This is because the end of the action is intrinsically linked to the start of the action.
%%
%That is, the start time and duration of the action will dictate the end time.

%\begin{figure}[h]
%  \centering
%  \includegraphics[width=0.8\linewidth]{figures/action_coordination}
%  \caption{Example showing the need to coordinate action start times.}
%  \label{fig: action coordination}
%\end{figure}

\subsubsection{Simple Temporal Network.}

An STN is a graph representation of a \textit{Simple Temporal Problem (STP)} \cite{dechter1991temporal} and is used to represent and reason about temporal information.
%
%Definition \ref{def: STN and bound of an edge} provides a formal definition of an $STN$.

\begin{definition}
\label{def: STN and bound of an edge}
A \textbf{Simple Temporal Network} is a directed edge-weighted graph $\langle V,E \rangle$ where $V$ is the set of nodes and $E$ the set of edges.
The nodes represent time-points and have continuous domains.
Each edge is labeled by a real value weight that is called the \textbf{bound} of the edge.
The bound of the edge from node $X$ to node $Y$ is denoted as $d_{XY}$ and the edge as $XY$.
The value assigned to a node $X$ is denoted as $T_X$.
Each edge imposes the following constraint on the values assigned to its end-points: $T_y - T_x \leq d_{XY}$.
%
% \begin{align}
% T_y - T_x \leq d_{XY} \notag
% \end{align}
%
In an STN no self-referencing edges or multiple edges (i.e. edges starting and ending at the same node) are allowed.
\end{definition}

\centerline{\includegraphics[width=0.3\linewidth]{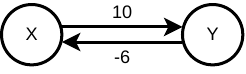}}
\noindent The above small STN with two nodes and two edges imposes the constraints: $6 \leq T_y - T_x \leq 10$, i.e., $Y$ must be between 6 and 10 units after $X$.

% Figure \ref{fig: simple temporal network} shows a small STN with two nodes and two edges.
% %
% Note that the two edges impose the following constraints:
% %
% \begin{align}
% 6 \leq T_y - T_x \leq 10 \notag
% \end{align}
% %
% The value of $Y$ must be between 6 and 10 units after $X$.

% \begin{figure}[h]
%   \centering
%   \includegraphics[width=0.3\linewidth]{figures/simple_temporal_network}
%   \caption{A Simple Temporal Network (STN).}
%   \label{fig: simple temporal network}
% \end{figure}

\subsection{Behavior Trees}
\label{ssec:bt_orig}

Behavior Trees (BTs) have become very popular in recent years as a means to encode robot control~\cite{6907656}.
%
% A BT is a hierarchical data structure defined recursively from a root node with several child nodes that, in turn, can have more children.
A BT is a hierarchical data structure defined recursively from a root node with several child nodes.

The basic operation of a BT node is the \textbf{tick}. When nodes are ticked, they can return three different values:
\begin{enumerate*}[label=\roman*)]
\item SUCCESS,
\item FAILURE, and
\item RUNNING.
\end{enumerate*}
%
%\begin{itemize}[itemsep=0pt,itemindent=0pt]
%  \item \textbf{SUCCESS}: It completed its mission successfully.
%  \item \textbf{FAILURE}: It failed in its mission.
%  \item \textbf{RUNNING}: It has not yet completed its mission.
%\end{itemize}
%
Furthermore, there are four different types of nodes:
\begin{itemize}[leftmargin=*]
\item \textbf{Control}: This type of node defines the execution flow of the tree. Examples of control nodes include sequence, fallback, and parallel nodes.
\item \textbf{Decorator}: This type of node can
\begin{enumerate*}[label=\roman*)]
\item transform the result received from the child,
\item halt the execution of the child, or
\item repeat the ticking of the child.
\end{enumerate*}
\item \textbf{Action}: This node is a leaf of the tree. An action node is implemented by the user to generate the low-level control required by the application.
\item \textbf{Condition}: This node returns SUCCESS to indicate the condition is met. Otherwise, it returns FAILURE.
\end{itemize}
\begin{figure}[tbh]
  \centering
  \includegraphics[width=\linewidth]{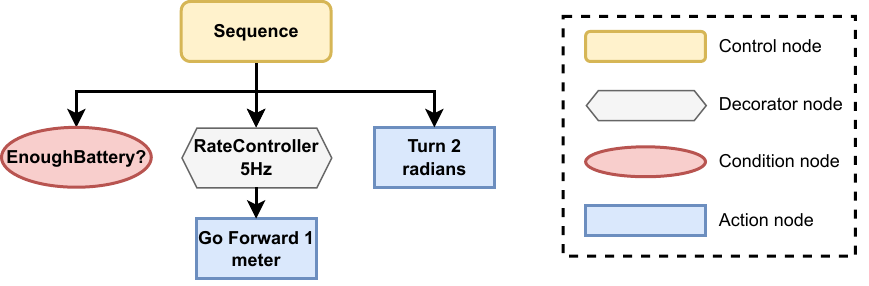}
  \caption{\label{img:bt_0}Simple Behavior Tree with various types of Nodes.}
\end{figure}
Figure \ref{img:bt_0} shows a simple BT.
When a BT is executed, the root node is ticked until it returns SUCCESS or FAILURE.
The root node is a sequence, so when it is ticked, it propagates the tick to its first child (\emph{EnougtBattery?}).
If this node returns SUCCESS (there is enough battery), the second child (\emph{RateController}) will then be ticked.
This decorator node controls the rate at which it ticks its children (\emph{Go Forward}).
\emph{Go Forward} will return RUNNING until the robot has moved one meter.
When \emph{Go Forward} and \emph{Turn} do their job and return SUCCESS, the sequence at the root will return SUCCESS, finishing the execution of the tree.

\section{Behavior Tree Algorithm}
\label{sec:bt_encoding}

Algorithm \ref{alg:GetBT} denotes the main algorithm of this paper and lists the three principal steps for transforming a temporal plan into an executable behavior tree.
The algorithm takes as input the temporal planning problem $TP$, the initial state $I$, the goal $R_g$, and the temporal plan $\pi$ solution of the planning problem.
In the first step, an STN is derived from the temporal plan, which encodes the explicit constraints imposed by the STP.
In the second step, the STN edge bounds are propagated to compute the bounds for every pair of nodes in the the STN.
Finally, in the third step, the BT is derived from the fully propagated STN.
%
%The following subsections provide the details of each of these steps.
In the following we provide the details of each of these steps.

%\todo[inline]{Josh: As Marco suggested, an alternative approach is to convert the STN into a ``dispatchable'' form. This would only require propagation to the neighbors of the time point to be executed.}

\begin{algorithm} [H]
\caption{GetBT Algorithm}
\label{alg:GetBT}
\begin{algorithmic}[1]
\Function{GetBT}{$TP,I,R_g,\pi$}
  \State $G \gets \Call{BuildSTN}{TP,I,R_g,\pi}$
  \State $G \gets \Call{Propagate}{G}$
  \State $BT \gets \Call{BuildBT}{G}$
  \State \Return $BT$
\EndFunction
\end{algorithmic}
\end{algorithm}

\subsection{Building the Graph}
\label{sec: Building the Graph}
The construction of the graph $G$ is decomposed in three steps:
\begin{enumerate*}[label=\roman*)]
\item derive an STN $G$ from the temporal plan $\pi$;
\item propagate the edge bounds to compute the bounds for every pair
  of nodes in $G$.
\end{enumerate*}
Hereafter we provide the essential details for
the construction of the graph $G$.

\subsubsection{Building the STN.}
\label{sec: Building the STN}

%The first step in Algorithm \ref{alg:GetBT} (\textsc{GetBT}) is to build the STN.
%
%See Algorithm \ref{alg:BuildSTN} (\textsc{BuildSTN}).
The construction of the STN is performed by \textsc{BuildSTN} whose pseudocode is illustrated in Algorithm \ref{alg:BuildSTN}.
The nodes of the STN are given by the snap actions of the actions in the temporal plan $\pi$.
%
% A snap action is an instantaneous action used to represent the start or end of a durative action.
%
Once the nodes have been created, explicit links are derived from the domain, plan, and initial state.
There are three types of explicit links:
\begin{enumerate*}[label=\roman*)]
\item \textit{\textbf{satisfying}}: created when an effect of one action satisfies a condition of another;
\item \textit{\textbf{threat avoidance}}:  created when the effects of one action would threaten the conditions or effects of another;
\item \textit{\textbf{temporal}}: created to encode the durational relationship between the start and end snap actions of a durative action.
\end{enumerate*}
These links are derived by processing the actions of the induced simple plan one at a time.
At each iteration, a search is performed to connect the current action to one or more earlier actions.
Actions are then traced to nodes in the graph to form links.
\begin{definition}
\label{def: graph node}
A \textbf{graph node} $n$ defines a start or end snap action corresponding to a durative action in the temporal plan and is composed of the following elements:
% \begin{itemize}[itemsep=0pt]
\begin{enumerate*}[label=$\bullet$)]
\item $n.\textbf{t}$: action start time,
\item $n.\textbf{a}$: action expression (name and parameters),
\item $n.\textbf{d}$: action duration,
\item $n.\textbf{l} \in \{ \textsc{start},\textsc{end} \}$: snap action type,
\item $n.\textbf{R}$: grounded conditions,
\item $n.\textbf{E}$: grounded effects,
\item $n.\textbf{U}$: input links,
\item $n.\textbf{Y}$: output links.
\end{enumerate*}
%\end{itemize}
Links are defined as list of tuples $\langle n,d_l,d_h \rangle$, where $n$ represents the parent or child node, and $d_l$,$d_h$  represents the lower and upper bounds of the start time of the child relative to the parent.
\end{definition}
For a graph $G$ we use \textsc{GetNodes}($a,G$) to retrieve the nodes in the graph $G$ corresponding to action $a$ if it exists.

\begin{algorithm} [tb]
\caption{BuildSTN Algorithm}
\label{alg:BuildSTN}
\begin{algorithmic}[1]
\Function{BuildSTN}{$TP,I,R_g,\pi$}
  \State $G \gets \Call{InitGraph}{TP,I,R_g,\pi}$
  \State $H \gets \Call{GetHappenings}{\pi}$
  \State $S \gets \Call{GetSimplePlan}{\pi}$
  \State $\{X\} \gets \Call{GetStates}{TP,I,H,S}$
  \For{$i \gets 1$ \textbf{to} $|S|$}
    \State $N_s \gets \Call{GetSatisfy}{S[i],TP,H,S,\{X\}}$
    \State $N_t \gets \Call{GetThreat}{S[i],TP,H,S,\{X\}}$
    \ForAll{$a_k \in \{N_s \cap N_t\}$}
      \ForAll{$n \in \Call{GetNodes}{S[i],G}$}
        \ForAll{$h \in \Call{GetNodes}{a_k,G}$}
          \State $G \gets \Call{PruneLinks}{n,h}$
          \If{$\neg \Call{CheckPaths}{n,h}$}
            \State $h.Y \gets \langle n,0,\infty \rangle$
            \State $n.U \gets \langle h,0,\infty \rangle$
          \EndIf
        \EndFor
      \EndFor
    \EndFor
  \EndFor
  \State \Return $G$
\EndFunction
\end{algorithmic}
\end{algorithm}

% The links of the STN are derived by processing the actions of the induced simple plan one at a time.
% %
% At each iteration, a search is performed to connect the current action to one or more earlier actions.
% %
% Actions are then traced to nodes in the graph to form links.

% \begin{itemize}[nosep]
At \textbf{line 2:} we initialize the STN by creating two nodes for each actions $n \in DA$ in the temporal plan $\pi$: $\astart{n}$ and $\aend{n}$, corresponding to its start and end snap actions.
The input and output links for these nodes are initialized as
%\begin{align}
%n_\vdash.U &= \{\} & n_\vdash.Y &= \{ \langle n_\dashv,d,d \rangle \} \notag \\
%n_\dashv.U &= \{ \langle n_\vdash,d,d \rangle \} &n_\dashv.Y &= \{\}, \notag
%\end{align}
%
$n_\vdash.U = \{\}$,  $n_\vdash.Y = \{ \langle n_\dashv,d,d \rangle \}$,
$n_\dashv.U = \{ \langle n_\vdash,d,d \rangle \}$, $n_\dashv.Y = \{\}$,
where $d$ denotes the duration of the durative action $n$.
The above links thus specify the constraint
%
%\begin{align}
%d \leq t_\dashv - t_\vdash \leq d.
%\end{align}
%
$d \leq t_\dashv - t_\vdash \leq d$.
A root node $n_0$ is created with effects defined by the initial state, $n_0.E = I$, and a goal node $n_g$ is created with conditions defined by the goal conditions, $n_g.R = R_g$.
At \textbf{lines 3-4} we use Definition \ref{def: induced simple plan} to compute the happening time points $H$ and induced simple plan $S$ from the input temporal plan $\pi$.
An additional action $a_0$ is prepended to the front of $S$ to represent the initial state $I$.
We then compute at \textbf{line 5:} the state sequence $\{X\} = \{I, \ldots, X(t_i), \ldots, X_f\}$
%
% \begin{align}
% \{X\} = \{I, \ldots, X(t_i), \ldots, X_f\} \notag
% \end{align}
%
corresponding to the happening time points $H = \{t_i\}$, where $X(t_i)$ specifies the value of the state vector, i.e., the conjunction of the predicates and function values, at happening time point $t_i$.
The state sequence is easily obtained from the initial state $I$ by applying the effects of the snap actions in the induced simple plan one at time.
Note that for a valid plan, the final state $X_f$ should contain the goal conditions.
Then we iterate over the induced simple plan (\textbf{lines 6-15}) to compute and add the satisfying and threat avoidance links.
Note that, the iteration starts with the first action from the standard induced simple plan, not from the auxiliary action $a_0$ corresponding to the initial state.
At \textbf{lines 7-8} we Iterate backward from the current action $i$ of the induced simple plan to find all satisfying actions $N_s$ and all actions $N_t$ that threaten or are threatened by $i$ by calling \textsc{GetSatisfy} and \textsc{GetThreat} respectively.
% \item \textbf{Line 7:} Iterate backward from the current action $i$ of the induced simple plan to find all satisfying actions.
% %
% See subsection below for a description of the \textsc{GetSatisfy} Algorithm and Appendix \ref{app: GetSatisfy Algorithm} for the full implementation details.
% %
% \item \textbf{Line 8:} Iterate backward from the current action $i$ of the induced simple plan to find all actions that threaten or are threatened by $i$.
% %
% See subsection below for a description of the \textsc{GetThreat} Algorithm and Appendix \ref{app: GetThreat Algorithm} for the full implementation details.
%
At \textbf{lines 9-15} we iterate over the combined set of satisfying and threat avoidance actions (representing the set of potential parents to the current action $i$) to prune previously established links that are not needed.
Before creating a link from the returned parent node $b$ to the current node $c$, look for an existing link $\overline{ac}$ where $a$ is a parent of $b$, i.e., $\exists$ a path $a \rightarrow b$.
If such a link exists, remove it, since $\overline{bc}$ will supersede $\overline{ac}$ in determining the causal relationship.
% \item \textbf{Line 9:} Iterate over the combined set of satisfying and threat avoidance actions.
% %
% These actions represent the set of potential parents to the current action $i$.
% %
% \item \textbf{Line 10-11:} Iterate over the nodes in the STN defined by the current action $i$ and the potential parent $k$.
% %
% The \textsc{GetNodes} function returns the nodes in the STN corresponding to an action in the induced simple plan.
% %
% Note that an overall action in the induced simple plan will map to both a start node and an end node in the STN.
% %
% \item \textbf{Line 12:} Prune previously established links that are not needed.
% %
% Before creating a link from the returned parent node $b$ to the current node $c$, look for an existing link $\overline{ac}$ where $a$ is a parent of $b$, i.e., $\exists$ a path $a \rightarrow b$.
% %
% If such a link exists, remove it, since $\overline{bc}$ will supersede $\overline{ac}$ in determining the causal relationship.
%
At \textbf{lines 13-15} we check for an existing path $n \rightarrow h$ from the returned parent node $h$ to the current node $n$ before creating a new link $\overline{hn}$.
If such a path exists, it is not necessary to create the new link, since the previously established path $n \rightarrow h$ will supersede $\overline{nh}$ in determining the causal relationship.\todo{Josh please check proper replacement with $a$ $c$ with $n$ and $h$ used in the algorithm!}
If such a link does not exist, then we create links to and from the
current node $n$ and the parent node $h$.  Note that the bounds on the
links are equivalent to the constraint $0 \leq t_n - t_h \leq \infty$.
%
%\begin{align}
%0 \leq t_n - t_h \leq \infty. \notag
%\end{align}
%\end{itemize}
%\end{trivlist}

\subsubsection{GetSatisfy Algorithm.}
\label{sec: GetSatisfy Algorithm}

The \textsc{GetSatisfy} algorithm starts from an input action $a_i$ of the induced simple plan and steps backward from $a_i$ to find satisfying actions.
Action $a_k$ is said to be a satisfying action of $a_i$ if $a_k$ contains an effect that satisfies a condition of $a_i$.
Let $X(t_{k-1})$ be the state before the effects of $a_k$ are applied and $\hat{X}$ be the state produced by applying the effects of $a_k$ to $X(t_{k-1})$, i.e., $\hat{X} \xleftarrow{E_k} X(t_{k-1})$.
An effect of $a_k$ satisfies a condition of $a_i$ if a condition of $a_k$ does not hold for $X(t_{k-1})$ and does hold for $\hat{X}$.

Note that, in general an effect of one action can satisfy a condition of another so long as the happening time of the satisfying action is less than or equal to the happening time of the satisfied action.
To find all satisfying actions, the search thus begins with the actions occurring at the same happening time as the input action.
Once the immediate actions have been evaluated, the search continues by stepping backward through the plan to evaluate all previous actions.

Appendix \ref{app: GetSatisfy Algorithm} provides the full implementation details and line-by-line explanation of the \textsc{GetSatisfy} Algorithm.

\subsubsection{GetThreat Algorithm.}
\label{sec: GetThreat Algorithm}

The \textsc{GetThreat} algorithm starts from an input action $a_i$ of the induced simple plan and steps backward from $a_i$ to find all actions that would threaten $a_i$ or be threatened by $a_i$.
Action $a_k$ is said to be threatened by action $a_i$ if $a_i$ contains an effect that threatens a condition or effect of $a_k$.
Let $X(t_{k-1})$ be the state before the effects of $a_k$ are applied and $\hat{X}$ be the state produced by applying the effects of $a_i$ to $X(t_{k-1})$, i.e., $\hat{X} \xleftarrow{E_i} X(t_{k-1})$.
Similarly, let $\bar{X}$ be the state produced by applying the effects of $a_k$ to $X(t_{k-1})$, i.e., $\bar{X} \xleftarrow{E_k} X(t_{k-1})$.
An effect of $a_i$ threatens a condition of $a_k$ if the conditions of $a_i$ hold for $X(t_{k-1})$ and the conditions of $a_k$ do not hold for $\hat{X}$.
Likewise, an effect of $a_k$ threatens a condition of $a_i$ if the conditions of $a_i$ hold for $X(t_{k-1})$ and do not hold for $\bar{X}$.
Additionally, an effect of $a_i$ threatens an effect of $a_k$ if the effect of $a_i$ contradicts an effect of $a_k$ and vice versa.
PDDL 2.1 goes further by requiring that two actions cannot be executed concurrently if they modify the same effect.
This requirement is known as the ``no moving targets'' rule.

Before it was noted that an effect of one action can satisfy a condition of another at the same happening time.
This implies that an action $a$ at happening time $t_i$ may only be satisfied once the state $X(t_{i-1})$ has been updated by one or more actions $\{a_k | a_k \in S(t_i), a_k \neq a\}$.
We say that an action $a$ is applicable at happening time $t_i$ if there exists an intermediate state $X' \gets X(t_{i-1})$ that satisfies the conditions of $a$, where the intermediate state $X'$ is obtained by applying zero or more actions $a_k \in S(t_i)$ to the state $X(t_{i-1})$.
In general, the number of possible intermediate states for a given happening time $t$ is given by the number of subsets in the power set of $\{a_k | a_k \in S(t)\}$.

When testing whether an action could threaten or be threatened by another action at a happening time $t_i$, it must first be determined if the action is applicable at $t_i$.
In the worst case, this requires checking all intermediate states defined by the power set of $\{a_k | a_k \!\in\! S(t_i)\}$.
However, in many cases the action will be satisfied by the previous state $X(t_{i-1})$.
Even if the full power set must be searched, the computation time will generally be small owing to the typically small number of actions for a given happening.

Appendix \ref{app: GetThreat Algorithm} provides the full implementation details and line-by-line explanation of the \textsc{GetThreat} Algorithm.

\subsubsection{Propagating the Constraints.}
\label{sec: Propagating the Constraints}

To propagate the edge bounds to compute the bounds for every pair of nodes in the graph we apply the Floyd Warshall Algorithm to compute all-pairs, shortest-path (APSP).

% %
% This is equivalent to solving an all-pairs, shortest-path (APSP) problem.

% In the previous section, we described how to derive the explicit links between the nodes in the STN using the domain, plan, and initial state.
% %
% In this section, we now describe how to propagate the edge bounds to compute the bounds for every pair of nodes in the graph.
% %
% This is equivalent to solving an all-pairs, shortest-path (APSP) problem.

\begin{lemma}\label{lemma:lemma-stn}
Given a time triggered plan $\pi$, the STN $G \gets \textsc{BuildSTN}(D,I,R_g,\pi)$ represents the temporal relation and causal dependencies of plan $\pi$.
\end{lemma}
\begin{proof}
The proof is trivial given Algorithm~\ref{alg:BuildSTN}.
\end{proof}

\subsection{Building the Behavior Tree}
\label{sec: Building the Behavior Tree}

% The final step of Algorithm \ref{alg:GetBT} (\textsc{GetBT}) is to build the behavior tree (BT) from the STN.
% %
% See Algorithm \ref{alg:BuildBT} (\textit{BuildBT}).
% %
There are two principal challenges in converting an STN to a BT.
The first involves representing the actions in the BT framework, and the second involves representing the logical and temporal constraints.

\begin{figure*}
\centering
\begin{tabular}{c@{\hspace{10mm}}c}
\includegraphics[width=0.8\columnwidth]{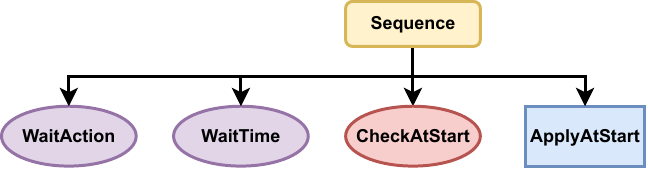} & \includegraphics[width=\columnwidth]{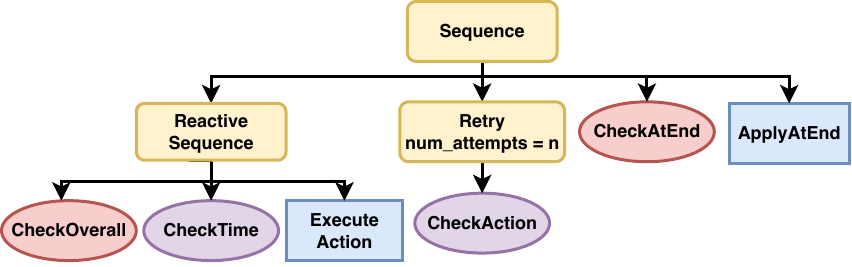}\\
\end{tabular}
\caption{Expansion of start action unit (left). Expansion of end action unit (right).}
  \label{fig: start action bt} \label{fig: end action bt}
\end{figure*}

BTs are hierarchical by design (see Sec.~\ref{ssec:bt_orig}):
a complex BT can be constructed from simpler BTs.
Similarly, a temporal plan $\pi$ is a combination of simple actions.
Taking inspiration from this parallelism, a time triggered plan can be converted to a BT by converting each durative action into a start and end BT and combing these BTs.
%
%Fig. \ref{fig: start action bt}  and Fig. \ref{fig: end action bt} represent respectively the start BT and the end BT templates.
Fig. \ref{fig: start action bt}  represent the start BT (left) and the end BT (right) templates.
The \textit{CheckAtStart}, \textit{CheckOverall}, and \textit{CheckAtEnd} nodes correspond to the \textit{at start}, \textit{overall}, and \textit{at end} conditions.
Likewise, the \textit{ApplyAtStart} and \textit{ApplyAtEnd} nodes correspond to the \textit{at start} and \textit{at end} effects.
A ``Check'' node prevents the flow from continuing if the specified conditions have not been met.
If the conditions have been met, an ``Apply'' node applies the action effects, thereby updating the state.
Finally, the \textit{ExecuteAction} node calls a service that executes the actual process represented by the action.

% \begin{figure}[tbh]
% \centering
% \includegraphics[width=0.85\linewidth]{figures/start_action_bt}
% \caption{Expansion of start action unit.}
% \label{fig: start action bt}
% \end{figure}
% %\todo[inline]{It would be good to have the images of the BT being parametric on what they are waiting, checking for, depending the action-node they are referring to.}

% \begin{figure}[tbh]
% \centering
% \includegraphics[width=\linewidth]{figures/end_action_bt}
% \caption{Expansion of end action unit.}
% \label{fig: end action bt}
% \end{figure}

The edges of an STN define the logical and temporal constraints governing the execution flow.
%
%A significant difference between an STN and a BT is that a node in an STN may have multiple parents, whereas a node in a BT can have only one.
%
%Therefore, to convert the STN to a BT, we must remove some of the parent links so that each node in the BT has at most one parent.
To convert the STN to a BT, we remove some of the parent links in the STN so that each node in the BT has at most one parent.
%
%The constraints imposed by the removed links are then enforced using \textit{WaitAction} and \textit{CheckAction} nodes, as shown in Figures \ref{fig: start action bt} and \ref{fig: end action bt}.
%
We enforce the constraints imposed by the removed links using \textit{WaitAction} and \textit{CheckAction} nodes, as shown in Figure \ref{fig: start action bt}.
A \textit{WaitAction} node causes the current branch of the BT execution to pause until the specified action has finished.
Likewise, a \textit{CheckAction} node causes the BT to fail if the specified action has not yet finished.
A \textit{WaitAction} node is used when the dependent node corresponds to a start action and a \textit{CheckAction} node is used when the dependent node corresponds to an end action.
%
%This is due to the fact that a start action can be paused, i.e., it can wait on another action to finish, whereas an end action cannot.
%
In general, an end action must start immediately after its start action, since the underlying durative action may represent a physical process with an end time dictated by the start time and the actual duration.
The \textit{WaitTime} node waits for a given amount of time and is used to enforce the start time of the associated action in the time triggered plan (or STN).
\textit{CheckTime} is used to prevent the execution if the specified time has already occurred.
\todo{Josh: It is not clear how the duration of the action is enforced@ Is it done by the ExecuteAction? We need to explain a bit this point!}
\todo{Josh: Please check if my interpretation is correct, and please fix it if wrong.}

To create a BT from an STN, we traverse the STN using a depth-first search (DFS), starting from the root node.
For each node, we look to see if the node has previously been \emph{used} by searching for the node in the used nodes set.
If the node has not been used, we add the node to the used nodes set, create the appropriate BT nodes, and continue to the next level of the STN.
If the node has been used, we create either a \textit{WaitAction} or a \textit{CheckAction} BT node, return to the previous level, and continue to the next output link.
In this way, no node is repeated in the BT and each node is guaranteed to have only one parent.
If the current STN node corresponds to a start action, the first child node to be visited is it's corresponding end action node.
This ensures that for each start action node in the BT there will be a corresponding end action node in the BT and the end action will be triggered immediately after the start action.

Algorithm~\ref{alg:BuildBT} illustrates the process. It takes in input the STN $G$ and returns a BT whose execution preserves the semantics of the plan $\pi$ that resulted in the STN $G$.
\begin{algorithm}[tb]
\caption{BuildBT Algorithm}
\label{alg:BuildBT}
%\todo[inline]{Marco: Here it is not completely clear what the different arguments do represent. In particular, the last one. Is $Q$ a set that is modified and propagated to the caller? Similarly, I suggest to remove the xml everywhere. The nodes of the graph do not seem to contain an id, thus the GetID what does it compute?}
\begin{algorithmic}[1]
\Function{BuildBT}{$G$}
  \State \Return $\Call{GetFlow}{n_0, \{\}, 1}$
%  \State $U \gets \{\}$
%  \State $BT \gets \Call{GetFlow}{n_0, U, 1}$
%  \State \Return $BT$
\EndFunction
\Statex
\Function{GetFlow}{$n,U,t$}
  \If{$n \in U$} \Return \Call{WaitAction}{$n,t$}
%    \State \Return \Call{WaitAction}{$n,t$}
  \EndIf
  \State $U \gets U \cup \{n\}$
  \If{$|n.Y| = 0$} \Return $\Call{ExecEnd}{n,t}$
%    \State \Return $\Call{ExecEnd}{n,t}$
  \EndIf
  \State $F \gets \{\}$
  \If{$n.l \neq \textsc{init}$} $F \gets F + \Call{SeqStart}{n,t}$
%    \State $F \gets F + \Call{SeqStart}{n,t}$
  \EndIf
  \If{$n.l = \textsc{start}$} $F \gets F + \Call{ExecStart}{n,t+1}$
%    \State $F \gets F + \Call{ExecStart}{n,t+1}$
  \ElsIf{$n.l = \textsc{end}$} $F \gets F + \Call{ExecEnd}{n,t+1}$
%    \State $F \gets F + \Call{ExecEnd}{n,t+1}$
  \EndIf
  \State $s \gets 0$
  \If{$|n.Y| > 1$}
    \State $F \gets F + \Call{ParStart}{|n.Y|,t+1}$
    \State $s \gets s + 1$
  \EndIf
  \If{$n.l = \textsc{start}$}
    \State $y \gets \Call{GetEnd}{n}$
    \State $F \gets F + \Call{GetFlow}{y,U,t+s+1}$
  \EndIf
  \ForAll{$y \in n.Y$}
    \If{$y \neq \Call{GetEnd}{n}$}
      \State $F \gets F + \Call{GetFlow}{y,U,t+s+1}$
    \EndIf
  \EndFor
  \If{$|n.Y| > 1$} $F \gets F + \Call{ParFinish}{t+1}$
%    \State $F \gets F + \Call{ParFinish}{t+1}$
  \EndIf
  \If{$n.l \neq \textsc{init}$} $F \gets F + \Call{SeqFinish}{t}$
%    \State $F \gets F + \Call{SeqFinish}{t}$
  \EndIf
  \State \Return $F$
\EndFunction
\end{algorithmic}
\end{algorithm}
\todo{Josh, please check the algorithm since the recursion before there where both y and n!}

\subsection{Correctness results}
The $BT_\pi \gets \textsc{GetBT}(DA, I, G, \pi)$ reflects the semantics of the time triggered plan $\pi$, and it if the plan $\pi$ is executable from $I$, then the execution of the BT from $I$ returns success. This is captured by the following theorem.

\begin{theorem}
  Given a temporal plan $\pi$ for a planning problem $(F, DA, I, G)$,
  the behavior tree $BT_\pi \gets \textsc{GetBT}(DA, I, G, \pi)$ when
  executed from a state $I$ returns SUCCESS iff the plan $\pi$ is
  executable in $I$. \todo{It is a first attempt, but we need to
    revise and/or add something more here.}
\end{theorem}
\begin{proof} (Sketch) The proof relies in the correct construction of the STN entailed from the time triggered plan $\pi$ (see Lemma~\ref{lemma:lemma-stn}). Indeed, in the construction of the STN $G$ we consider the time points related to the start and end of the actions in the plan, the action start and the duration of the actions as specified in the plan. Moreover, in the STN we also consider causal dependencies among the actions. Thus, the STN $G$ considers all the constraints governing the execution. Each start and end time point in the STN $G$ is encoded with a corresponding BT template instantiated with the action the node corresponds to. These templates checks the preconditions, apply the effects, and wait for the proper time thus capturing the semantics of the respective snap action and in turn of the respective durative action. The algorithm~\ref{alg:BuildBT} combines these instantiated templates in sequence/parallel BT nodes considering the structure of the STN induced by the plan. Thus, if the initial plan $\pi$ is executable from $I$, then the execution of the $BT_\pi$ returns success if executed from the same state $I$.
\end{proof}

\section{Experimental Validation}
\label{sec:validation}

In this section, we validate our approach analytically and experimentally in a real robot our contribution in this paper.

The baseline against which we will compare our approach is the previous version \cite{DBLP:conf/atal/RicoMELO21} of the Behavior
Tree Creator module in PlanSys2. The main difference is that it considers each plan action as a single node in the graph.
This approach has the disadvantage to our approach that it is incapable of modeling specific temporal dependencies,
such as when an action \emph{a} should be executed after the start of another action \emph{b}, but finish before b ends.
This effect is due to the inability to model dependencies in different phases of an action (start or end) but can only model
dependencies to an action atomically. The consequences are:
\begin{enumerate}
  \item Some plans cannot be executed.
  \item Some plans are executed sequentially when the planner indicates they must be executed in parallel.
\end{enumerate}

With the previous approach, PlanSys2 has been used successfully in multiple projects by many organizations, such as the
US Navy or NASA. However, they have either avoided this situation or assumed an inevitable parallelism loss. Our contribution
aims to fix these limitations.

Taking the widely known MATCHCELLAR domain \cite{LINARESLOPEZ201582}, let's discuss why our approach can deal with plans that
the baseline os not able to deal with. For a simple problem, we would produce the next plan:

% \tiny
% \begin{verbatimtab}
%   (define (domain matchcellar)
%   (:requirements :durative-actions :typing)
%   (:types match fuse - object)
%
%   (:predicates
%     (light ?m - match)
%     (handfree)
%     (unused ?m - match)
%     (mended ?f - fuse)
%   )
%
%   (:durative-action LIGHT_MATCH
%     :parameters (?m - match)
%     :duration (= ?duration 8)
%     :condition (and (at start (unused ?m))
%                     (over all (light ?m)))
%     :effect (and (at start (not (unused ?m)))
%                  (at start (light ?m))
%                  (at end (not (light ?m))))
%   )
%
%   (:durative-action MEND_FUSE
%     :parameters (?f - fuse ?m - match)
%     :duration (= ?duration 5)
%     :condition (and (at start (handfree))
%                     (over all (light ?m)))
%     :effect (and (at start (not (handfree)))
%                  (at end (mended ?f))
%                  (at end (handfree)))
%   )
% )
% \end{verbatimtab}
% \normalsize
%
% A simple problem like the next,
%
% \tiny
% \begin{verbatimtab}
% (define (problem fixfuse)
%   (:domain matchcellar)
%
%   (:objects
%     match1 match2 - match
%     fuse1 fuse2 - fuse
%   )
%
%   (:init
%     (unused match1)
%     (unused match2)
%     (handfree)
%   )
%
%   (:goal
%     (and
%       (mended fuse1)
%       (mended fuse2)
%     )
%   )
%
%   (:metric minimize (total-time))
% )
% \end{verbatimtab}
% \normalsize
%

\tiny
\begin{verbatimtab}
0.000: (light_match match1)  [8.000]
0.001: (mend_fuse fuse1 match1)  [5.000]
2.002: (light_match match2)  [8.000]
5.002: (mend_fuse fuse2 match2)  [5.000]
\end{verbatimtab}
\normalsize

\begin{figure}[tbh]
  \centering
  \includegraphics[width=\linewidth]{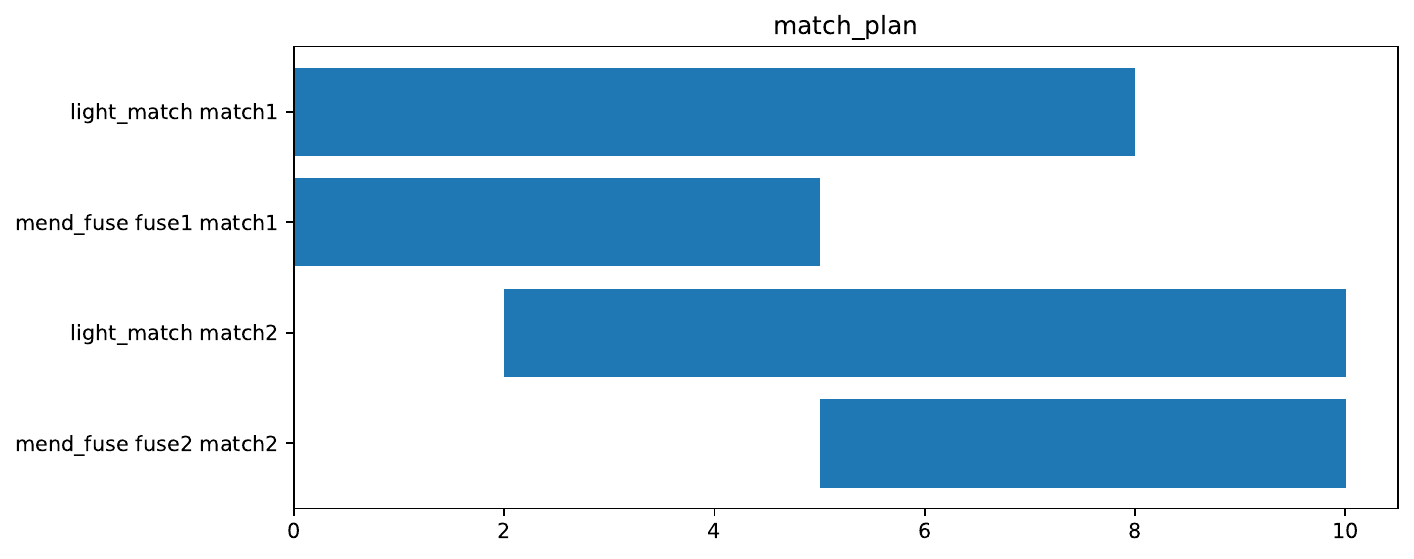}
  \caption{Gantt Chart of the plan generated to solve the MATCHCELLAR problem.}
  \label{fig:match}
\end{figure}

Figure \ref{fig:match} shows the execution Gantt Chart of the previous plan. Both actions \texttt{(light\_match match1)} and
\texttt{(light\_match match2)} can be executed in parallel, as they are independent. \texttt{(mend\_fuse fuse1 match1)} and
\texttt{(mend\_fuse fuse2 match2)} can not be executed in parallel, as they depend on the predicate\texttt{(handfree)}, and each
one removes it at the start and restores at the end. It is similar to thinking, "each of these actions needs to acquire the
resource \texttt{(handfree)} to execute." Furthermore, the action \texttt{(mend\_fuse fuse1 match1)} can be executed only
while \texttt{(light\_match match1)} is executing because it needs a predicate \texttt{(light match1)} all the time it is
running, and this predicate is added at the start of \texttt{(light\_match match1)}, and removed at its end. The same happened
with the action \texttt{(mend\_fuse fuse2 match2)} can be executed only between the start of \texttt{(light\_match match2)}.

The previous version (the baseline) of the Behavior Tree creator transforms a plan into a behavior tree, starting by creating a
dependency graph. If this graph cannot be created, the resulting Behavior Tree cannot be created, and the plan cannot be executed.
This version treated each action atomically. This version was unable  to code that action \texttt{(mend\_fuse fuse1 match1)}
should be executed during action \texttt{(light\_match match1)} since it added and removed a requirement of \texttt{(mend\_fuse fuse1 match1)}.
This version representation could not establish an order of execution between the two, knowing that they were dependent.
On the contrary, this approach established that \texttt{(light\_match match1)} and \texttt{(mend\_fuse fuse1 match1)} should be
executed one after the other since they were incompatible.

Our contribution divides each action into phases in this graph, being able to determine the order of execution of the actions
since it codifies that \texttt{(mend\_fuse fuse1 match1)} must be executed between the beginning and the end of \texttt{(light\_match match1)}.
 So, our approach can solve the execution of this plan.

%, as can be seen in Figure \ref{fig:matchgraph1}.

% \begin{figure}[tbh]
%   \centering
%   \includegraphics[width=\linewidth]{figures/MatchCellar2.pdf}
%   \caption{Dependency graph for MATCHCELLAR plan, consider actions atomically (baseline).}
%   \label{fig:matchgraph2}
% \end{figure}
%
% \begin{figure}[tbh]
%   \centering
%   \includegraphics[width=0.8\linewidth]{figures/MatchCellar1.pdf}
%   \caption{Dependency graph for MATCHCELLAR plan, dividing actions in phases (our contribution).}
%   \label{fig:matchgraph1}
% \end{figure}

The following experiment shows how our approach allows for optimizing the execution of a plan to
the baseline. We propose an experiment in which a robot must assemble cars in a factory. The car parts
are in different zones (body car, wheels, and steering wheels zones) and are assembled in an assembly zone.
The robot must move between zones to transport the parts to the assembly zone. In this case, the robot divides
the handling of a part into a pre-pick phase to put the arm in position and a pick phase to pick up the part.
The same happens with the release action, executed after a pre-release action. Pick and release can only be
executed at the designated position within a zone, but pre-pick and pre-release actions can be executed while the
robot reaches each zone.

\begin{figure}[tbh]
  \centering
  \includegraphics[width=0.8\linewidth]{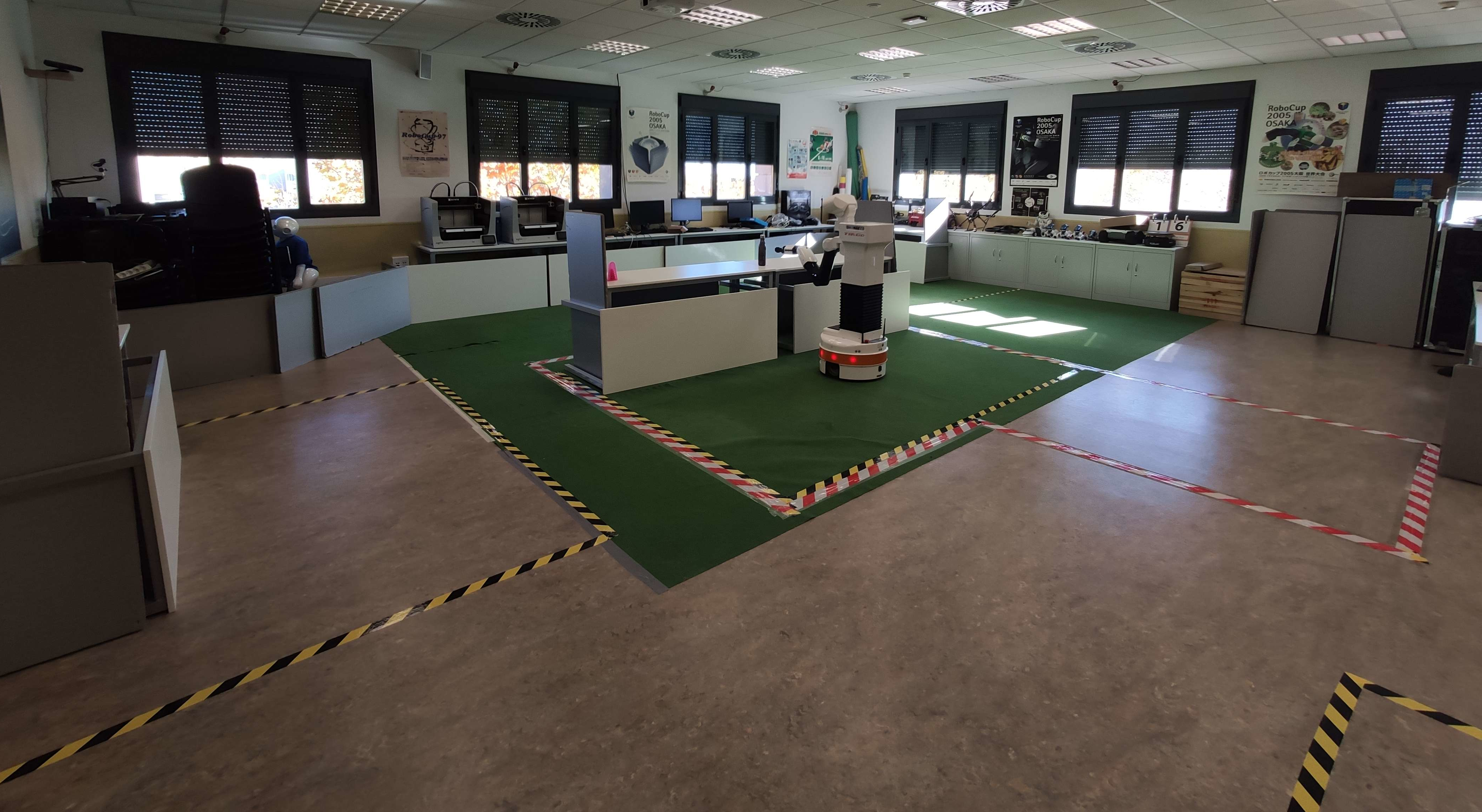}
  \caption{Experiment setup. Tiago robot and a laboratory with marked zones.}
  \label{fig:setup}
\end{figure}

\begin{figure}[tbh]
  \centering
  \includegraphics[width=0.6\linewidth]{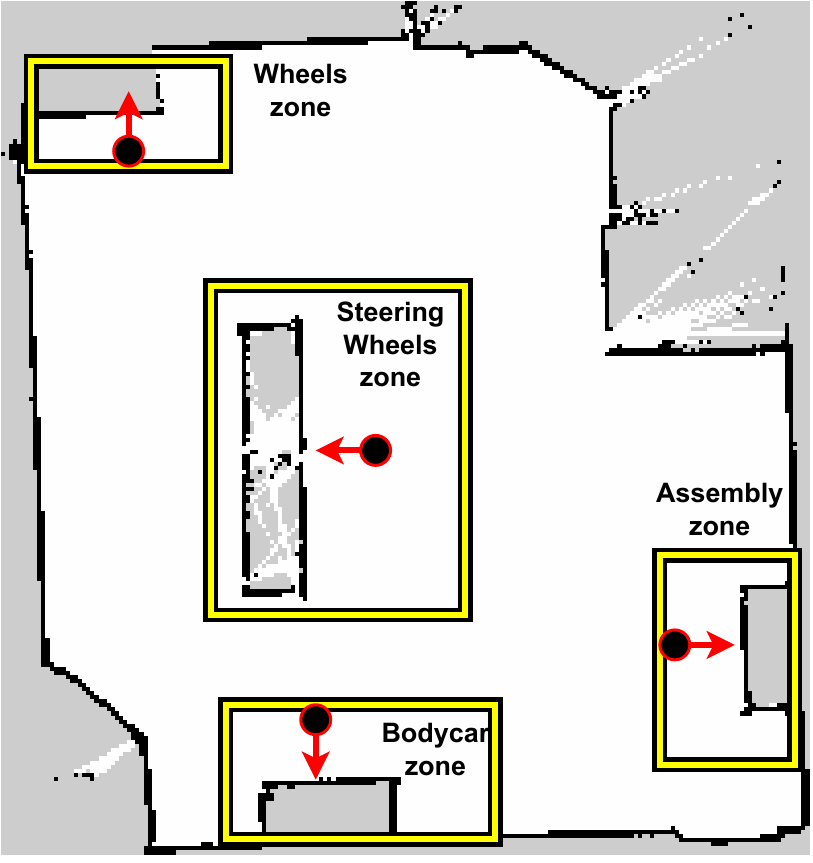}
  \caption{Experiment setup. Tiago robot and a laboratory with marked zones.}
  \label{fig:setup2}
\end{figure}

We carry out this experiment on the real Tiago robot in our laboratory (Figure \ref{fig:setup}),
simulating this industrial assembly environment as shown in Figure \ref{fig:setup2}, where four zones are
delimited. Each zone has an associated waypoint to which the robot must navigate to be in the operation point
of the zone.

We have used PlanSys2 to execute the plan in the real robot, commanding navigation actions to Nav2 and
joint positions to the arm. In the experiment, the robot carries out a mission whose plan is as follows:

\tiny
\begin{verbatimtab}
A1  0.000: (move r2d2 assembly_zone body_car_zone)  [20.000]
A2  15.001: (prepick r2d2 body_car_1 body_car_zone)  [5.000]
A3  20.002: (pick r2d2 body_car_1 body_car_zone)  [5.000]
A4  25.002: (move r2d2 body_car_zone assembly_zone)  [20.000]
A5  40.003: (prerelease r2d2 body_car_1 assembly_zone)  [5.000]
A6  45.004: (release r2d2 body_car_1 assembly_zone)  [5.000]
A7  50.004: (move r2d2 assembly_zone steering_wheels_zone)  [20.000]
A8  65.005: (prepick r2d2 steering_wheel_1 steering_wheels_zone)  [5.000]
A9  70.006: (pick r2d2 steering_wheel_1 steering_wheels_zone)  [5.000]
A10 75.006: (move r2d2 steering_wheels_zone assembly_zone)  [20.000]
A11 90.007: (prerelease r2d2 steering_wheel_1 assembly_zone)  [5.000]
A12 95.008: (release r2d2 steering_wheel_1 assembly_zone)  [5.000]
A13 100.008: (move r2d2 assembly_zone wheels_zone)  [20.000]
A14 115.009: (prepick r2d2 wheel_1 wheels_zone)  [5.000]
A15 120.010: (pick r2d2 wheel_1 wheels_zone)  [5.000]
A16 125.010: (move r2d2 wheels_zone assembly_zone)  [20.000]
A17 140.011: (prerelease r2d2 wheel_1 assembly_zone)  [5.000]
A18 145.012: (release r2d2 wheel_1 assembly_zone)  [5.000]

\end{verbatimtab}
\normalsize

When executing this plan on a real robot, the times of the action that include moving the arm can be
known a priori, but the navigation times have a significant variance. The distances between waypoints are
different, and the robot can vary its stopping times, for example.

\begin{figure}[tbh]
  \centering
  \includegraphics[width=\linewidth]{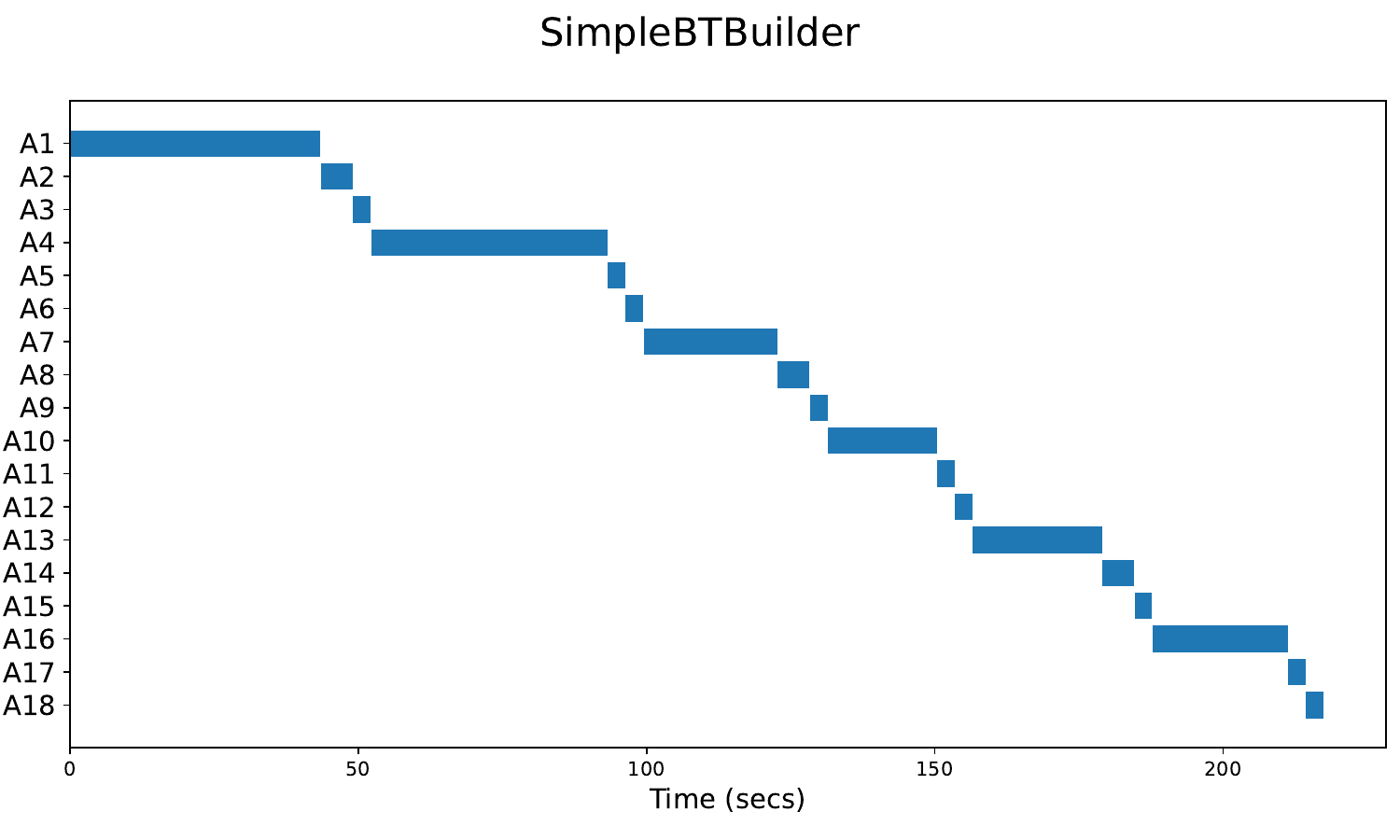}
  \caption{One iteration with SimpleBTBuilder.}
  \label{fig:res_simple}
\end{figure}
\begin{figure}[tbh]
  \centering
  \includegraphics[width=\linewidth]{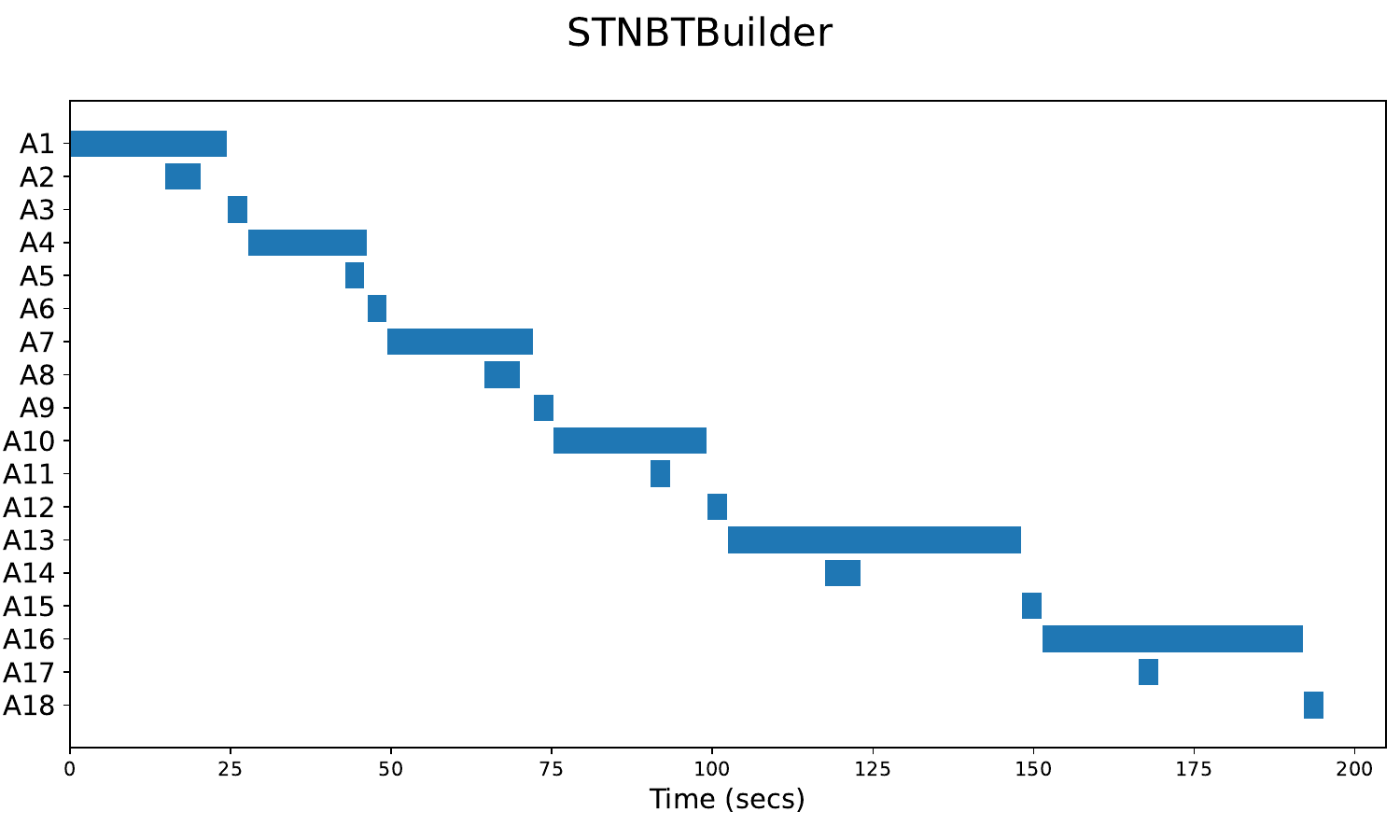}
  \caption{One iteration with STNBTBuilder.}
  \label{fig:res_simple}
\end{figure}

In this experiment, we ran the plan ten times with the baseline (SimpleBTBuilder) and ten times with our approach
(STNBTBuilder). The metric is the total execution time of the plan. As shown in Figure \ref{fig:res_simple}, with the previous approach,
although the plan indicates that actions must be executed in parallel, all actions are executed sequentially. With our
approach, it can be seen in Figure \ref{fig:res_stn} how the actions planned to be executed in parallel overlap their execution, improving
the execution time of the plan.

\scriptsize
\begin{table}[tbh]
  \begin{tabular}{|l|l|l|l|l|l|}
  \hline
                           & \textbf{Mean} & \textbf{Stdev} & \textbf{Median} & \textbf{Max} & \textbf{Min} \\ \hline
  \textbf{Simple} & 220.57                     & 9.21            & 216.37              & 245.80            & 215.79            \\ \hline
  \textbf{STN}    & 200.20                     & 10.92           & 195.09              & 222.49            & 191.49            \\ \hline
  \end{tabular}
\caption{Execution times (seconds) for each algorithm.}\label{tab:exp2}
\end{table}
\normalsize
Table \ref{tab:exp2} shows how our approach improves the average execution time to the baseline by respecting the planned parallelization of the plan's actions.

%\subsection{Limitations}

\todo[inline]{Fmrico: We call dedicate one-two paragraph to present any limitation.}

\section{Conclusions and future work}
\label{sec:conclusions}

\todo[inline]{Fmrico: Conclusions of our work.}

Executing temporal plans in the real and open world requires adapting to uncertainty both in the environment and in the plan actions.
A plan executor must therefore be flexible to dispatch actions based on the actual execution conditions.
In general, this involves considering both event and time-based constraints between the actions in the plan.
A simple temporal network (STN) is a convenient framework for specifying the constraints between actions in the plan.
Likewise, a behavior tree (BT) is a convenient framework for controlling the execution flow of the actions in the plan.
In this paper we presented an algorithm for transforming a plan into
an STN, and an algorithm for transforming an STN into a BT, thus a
systematic approach for executing total-order (time-triggered) plans
in robots operating in the real world.
The approach first creates a graph describing a deordered (state-triggered) plan and then creates a BT representing a partial-order (determined at runtime) plan.
The proposed algorithms ensure the correct execution of plans, including those with required concurrency.
We demonstrated the validity of our approach within the PlanSys2
framework on real robots.

As future work we aim to integrate notions of robust executions based
on the computation of robust envelopes and the integration of online
planning to enable for intertwined planning and execution.

\iffalse
\section{Acknowledgments}

\todo[inline]{Josh: Add thank you to all PlanSys2 contributors and projects, if any.}

M. Roveri is partially funded by the project MUR PRIN 2020 - RIPER - Resilient AI-Based Self-Programming and Strategic Reasoning - CUP  E63C22000400001.
This work is partially funded by the AIPlan4EU H2020 project (https://aiplan4eu-project.eu) - European Commission under grant agreement number 101016442 and
the DMARCE project - Ministerio de Ciencia e Innovación (Grant N. PID2021-126592OB-C22).

\fi
\clearpage

\bibliographystyle{IEEEtran}
\bibliography{Bibliography-File}

\clearpage
\appendix

\section{GetSatisfy Algorithm}
\label{app: GetSatisfy Algorithm}

The following pseudocode provides the implementation details for the \textsc{GetSatisfy} Algorithm previously described in Section \ref{sec: GetSatisfy Algorithm}.

\begin{algorithm} [H]
\caption{GetSatisfy Algorithm}
\label{alg:GetSatisfy}
\begin{algorithmic}[1]
\Function{GetSatisfy}{$a,D,H,S,\{X\}$}
  \State $N \gets \{\}$
  \State $t_2 \gets \Call{GetHappening}{a.t,H}$
  \State $R_a \gets \Call{GetConds}{a,D}$
  \While{$t_2 \geq 0$}
    \State $t_1 \gets \Call{GetPrevious}{t_2,H}$
    \State $X_1 \gets \Call{GetState}{t_1,\{X\}}$
    \ForAll{$r \in R_a$}
      \If{$\neg \Call{Check}{r,X_1}$}
        \ForAll{$a_k \in S(t_2)$}
          \If{$a_k = a$}
            \State \Call{Continue}{}
          \EndIf
          \State $E_k \gets \Call{GetEffs}{a_k,D}$
          \State $\hat{X} \gets \Call{Apply}{E_k,X_1}$
          \If{$\Call{Check}{r,\hat{X}}$}
            \State $N \gets a_k$
          \EndIf
        \EndFor
      \EndIf
    \EndFor
    \State $t_2 \gets t_1$
  \EndWhile
  \ForAll{$r \in R_a$}
    \If{\Call{Check}{$r,I$}}
      \State $N \gets a_0$
    \EndIf
  \EndFor
  \State \Return $N$
\EndFunction
\end{algorithmic}
\end{algorithm}

\begin{itemize}
\item \textbf{Line 3:} Get the happening time $t_2$ for the input action $a \in S$.
If $a$ is of type \textsc{overall}, $t_2$ is the last happening time such that $t_2 < a.t$.
\item \textbf{Line 4:} Get the grounded conditions $R_a$ of the input action $a \in S$.
\item \textbf{Line 5:} Step backward through the happening time points while the current happening time $t_2$ is greater than or equal to zero.
\item \textbf{Line 6:} Get the happening time $t_1$ prior to $t_2$.
\item \textbf{Line 7:} Get the state vector $X_1 = X(t_1)$.
For the first iteration, $X_1$ is the state prior to the application of all actions $\{a_k | a_k \in S(t_i)\}$, where $t_i = t_1$ is the happening time associated with the input action.
For the second iteration, $X_1$ is the state prior to the application of all actions $\{a_k | a_k \in S(t_{i-1})\}$, where $t_{i-1}$ is the previous happening time, and so on.
\item \textbf{Line 8:} Iterate over the grounded conditions $R_a$ of the input action $a$.
\item \textbf{Line 9:} Check if the condition $r \in R_a$ of the input action $a$ is satisfied by $X_1$.
If the condition is not satisfied, proceed to see if there exists an action $a_k \in S(t_2)$ that satisfies it.
\item \textbf{Line 10:} Iterate over the actions $a_k \in S(t_2)$, i.e., the actions occurring at happening time $t_2$.
\item \textbf{Lines 11-13:} Skip the input action.
An action cannot satisfy itself.
\item \textbf{Lines 14-15:} Apply the effects $E_k$ of action $a_k$ to $X_1$ to get $\hat{X}$, i.e., $\hat{X} \xleftarrow{E_k} X_1$.
\item \textbf{Lines 16-18:} Check if the condition $r \in R_a$ of the input action $a$ is satisfied by $\hat{X}$.
If so, add $a_k$ to the return set.
Action $a_k$ is a satisfying action.
\item \textbf{Line 22:} Set $t_2$ to $t_1$ to move the search back one happening time point.
\item \textbf{Lines 24-28:} Once all prior actions have been tested, test condition $r \in R_a$ against the initial conditions $I$.
If an initial condition $x \in I$ satisfies $r$, add $a_0$ to the return set.
\end{itemize}

\section{GetThreat Algorithm}
\label{app: GetThreat Algorithm}

The following pseudocode provides the implementation details for the \textsc{GetThreat} Algorithm previously described in Section \ref{sec: GetThreat Algorithm}.

\begin{algorithm} [H]
\caption{GetThreat Algorithm}
\label{alg:GetThreat}
\begin{algorithmic}[1]
\Function{GetThreat}{$a,D,H,S,\{X\}$}
  \State $N \gets \{\}$
  \State $t_2 \gets \Call{GetHappening}{a.t,H}$
  \State $R_a \gets \Call{GetConds}{a,D}$
  \State $E_a \gets \Call{GetEffs}{a,D}$
  \While{$t_2 \geq 0$}
    \State $t_1 \gets \Call{GetPrevious}{t_2,H}$
    \State $X_1 \gets \Call{GetState}{t_1,\{X\}}$
    \State $X_1^a \gets \Call{GetInter}{a,S,t_2,X_1}$
    \If{\Call{Check}{$R_a,X_1^a$}}
      \State $t_3 \gets \Call{GetNext}{t_2,H}$
      \ForAll{$a_k \in S(t)\ \forall t \in [t_2,t_3)$}
        \If{$a_k = a$}
          \State \Call{Continue}{}
        \EndIf
        \State $R_k \gets \Call{GetConds}{a_k,D}$
        \State $E_k \gets \Call{GetEffs}{a_k,D}$
        \State $X_1^k \gets \Call{GetInter}{a_k,S,t_2,X_1}$
        \State $\hat{X} \gets \Call{Apply}{E_a,X_1^k}$
        \State $\bar{X} \gets \Call{Apply}{E_k,X_1^a}$
        \If{$\neg \Call{IsOverall}{a} \land$ \\ $\neg \Call{Check}{R_k,\hat{X}}$}
          \State $N \gets a_k$
        \ElsIf{$\neg \Call{IsOverall}{a_k} \land$ \\ $\neg \Call{Check}{R_a,\bar{X}}$}
          \State $N \gets a_k$
        \ElsIf{$\neg \Call{IsOverall}{a} \land$ \\ $\neg \Call{IsOverall}{a_k}$}
          \State $d\hat{X} = \Call{Diff}{X_1^k,\hat{X}}$
          \State $d\bar{X} = \Call{Diff}{X_1^a,\bar{X}}$
          \If{$d\hat{X} \cap d\bar{X} \neq \{\}$}
            \State $N \gets a_k$
          \EndIf
        \EndIf
      \EndFor
    \EndIf
    \State $t_2 \gets t_1$
  \EndWhile
  \State \Return $N$
\EndFunction
\end{algorithmic}
\end{algorithm}

\begin{itemize}
\item \textbf{Line 3:} Get the happening time $t_2$ for the input action $a \in S$.
If $a$ is of type \textsc{overall}, $t_2$ is the last happening time such that $t_2 < a.t$.
\item \textbf{Lines 4-5:} Get the grounded conditions $R_a$ and effects $E_a$ of the input action $a \in S$.
\item \textbf{Line 6:} Step backward through the happening time points while the current happening time $t_2$ is greater than or equal to zero.
\item \textbf{Line 7:} Get the happening time $t_1$ prior to $t_2$.
\item \textbf{Line 8:} Get the state vector $X_1 = X(t_1)$.
For the first iteration, $X_1$ is the state prior to the application of all actions $\{a_k | a_k \in S(t_i)\}$, where $t_i = t_1$ is the happening time associated with the input action.
For the second iteration, $X_1$ is the state prior to the application of all actions $\{a_k | a_k \in S(t_{i-1})\}$, where $t_{i-1}$ is the previous happening time, and so on.
\item \textbf{Line 9:} Get the first intermediate state $X_1^a$ that satisfies the conditions $R_a$ of the input action.
Recall that $X_1$ represents the state prior to applying the effects of the actions in the happening set $\{a_k | a_k \in S(t_2)\}$.
The intermediate states are then obtained by applying zero or more actions from $\{a_k | a_k \in S(t_2)\}$ to $X_1$.
The set of all possible intermediate states is given by the power set of $\{a_k | a_k \in S(t_2)\}$.
If no suitable intermediate state can be found, return $X_1$.
\item \textbf{Line 10:} Check if the conditions $R_a$ of the input action $a$ are satisfied by $X_1^a$.
\item \textbf{Line 11:} Get the happening time $t_3$ after $t_2$.
\item \textbf{Line 12:} Iterate over the actions in $\{a_k | a_k \in S(t)\ \forall\ t_2 \leq t < t_3\}$. If the previous check passed, then one or more of these actions could threaten or be threatened by $a$.
\item \textbf{Lines 13-15:} Skip the input action.
An action cannot threaten or be threatened by itself.
\item \textbf{Lines 16-17:} Get the grounded conditions $R_k$ and effects $E_k$ of action $a_k \in S(t)$, where $t_2 \leq t < t_3$.
\item \textbf{Line 18:} Get the first intermediate state $X_1^k$ that satisfies the conditions $R_k$ of action $a_k$.
Such a state is guaranteed to exist if the plan is valid.
\item \textbf{Line 19:} Apply the effects $E_a$ of the input action $a$ to state $X_1^a$ to get $\hat{X}$, i.e., $\hat{X} \xleftarrow{E_a} X_1^a$.
\item \textbf{Lines 21-22:} If the input action is not of type \textsc{overall} and the conditions $R_k$ of action $a_k$ are not satisfied by $\hat{X}$, add $a_k$ to the return set.
The input action threatens action $a_k$.
\item \textbf{Lines 23-24:} If action $a_k$ is not of type \textsc{overall} and the conditions $R_a$ of the input action $a$ are not satisfied by $\bar{X}$, add $a_k$ to the return set.
The input action is threatened by action $a_k$.
\item \textbf{Line 25:} Check if the input action $a$ or the test action $a_k$ is of type \textsc{overall}.
If neither is of type \textsc{overall}, continue to check if the ``no moving targets'' rule will be violated.
\item \textbf{Line 26-27:} Compute the set difference $d\hat{X}$ between $X_1^k$ and $\hat{X}$ and the set difference $d\bar{X}$ between $X_1^a$ and $\bar{X}$.
$d\hat{X}$ denotes the set of states affected by the input action $a$.
$d\bar{X}$ denotes the set of states affected by the test action $a_k$.
\item \textbf{Lines 28:} Compute the intersection of the set differences $d\hat{X} \cap d\bar{X}$.
If the intersection is non-empty, add $a_k$ to the return set.
The input action $a$ and the test action $a_k$ modify the same effect, violating the ``no moving targets'' rule.
\item \textbf{Line 34:} Set $t_2$ to $t_1$ to move the search back one happening time point.
\end{itemize}

\section{The PlanSys2 ROS 2 Planning Framework}
\todo{We may skip this section since it seems not needed for the paper}
PlanSys2~\cite{PlanSys2} is a ROS 2~\cite{doi:10.1126/scirobotics.abm6074} Framework that allows executing plans generated by a PDDL planner in one or more robots. The general architecture is shown in Figure~\ref{fig:plansys2}, with a modular design composed of the following elements:

\begin{figure}[h]
  \centering
  \includegraphics[width=\linewidth]{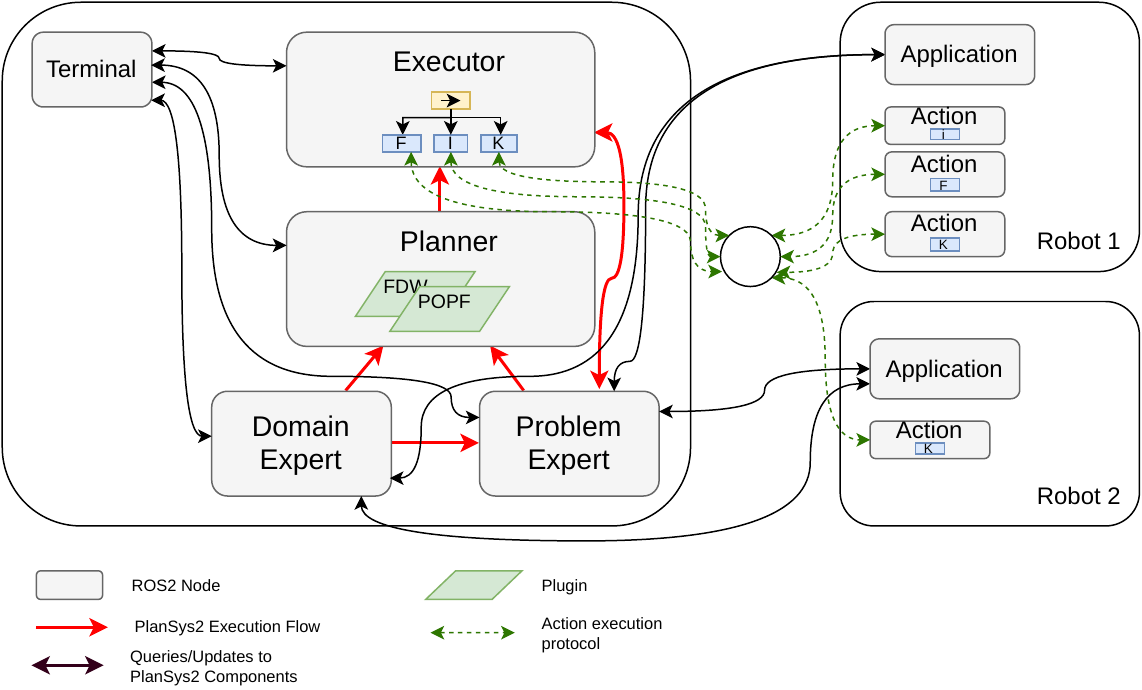}
  \caption{PlanSys2 Architecture.}
  \label{fig:plansys2}
\end{figure}

\begin{itemize}
  \item \textbf{Domain Expert}: It is responsible for reading the PDDL domain and responding to queries referring to
  this domain from other modules. It is fundamentally static.
  \item \textbf{Problem Expert}: Stores the PDDL problem. It can be accessed to add/remove instances, predicates,
  fluents, and, in general, any problem element.
  \item \textbf{Planner}: The module executes a PDDL planner with the content of the two previous modules. It is
  possible to use different planners by implementing a plugin with the command that runs the planner program and parses the output.
  \item \textbf{Executor}: Executes a plan calculated by the previous module.
  \item \textbf{Tools}: PlanSys2 has some tools to help the Roboticist to monitor the plan execution and interact with the system.
  The \emph{terminal} is a command line application that allows interacting with all modules to query the domain, update the problem, ask for a plan, or run an execution, among others. Many tools graphically display the status of the action performers, the problem content, and the execution process.
  \item \textbf{User application}: User applications are composed of a PDDL domain, the implementation of each of the domain actions,
  following an interface provided by PlanSys2, and (optionally) a controller program that initiates the knowledge of the Problem Expert
  and requests when executing it executes a plan, controlling its result.
\end{itemize}

\begin{figure*}[tb]
  \centering
  \includegraphics[width=0.4\linewidth]{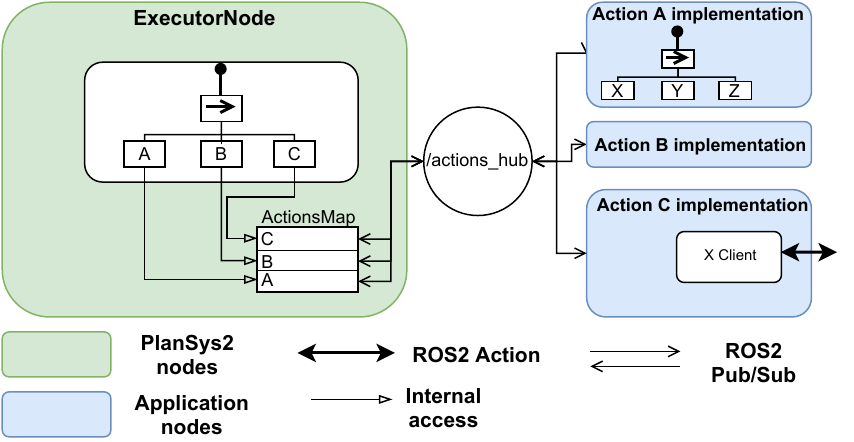}
  \includegraphics[width=0.58\linewidth]{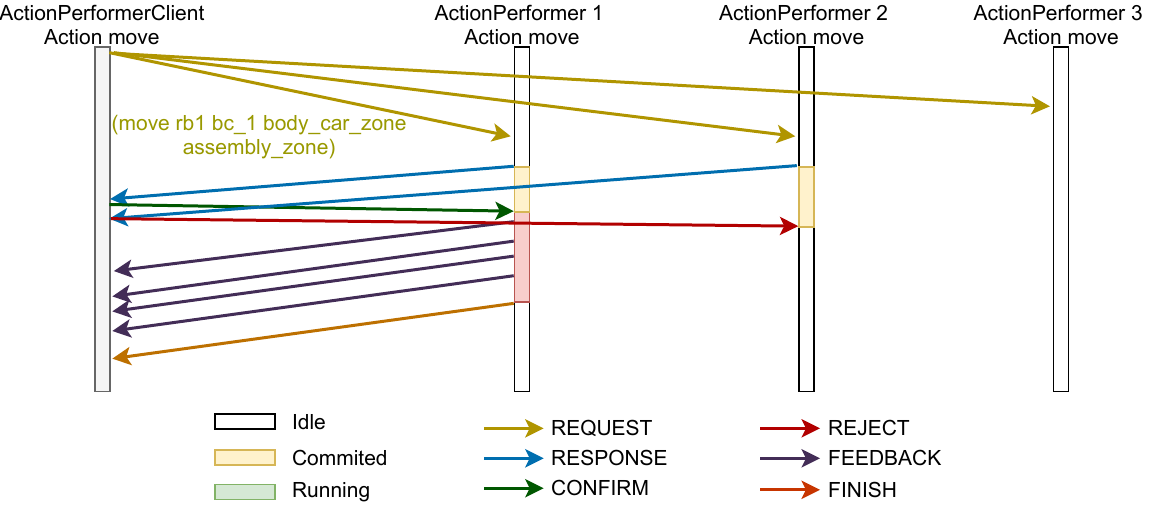}
  \caption{PlanSys2 execution protocol.}
  \label{fig:protocol}
\end{figure*}

The most relevant module for this work is the Executor. This module starts from a plan, builds the computation graph,
and converts it to a behavior tree ready to run. During the execution of the plan, it uses a protocol (Figure \ref{fig:protocol}) to request the action
performers to execute an action, receiving feedback and controlling its completion and result.

\end{document}